%% file: main.tex
\documentclass[runningheads]{llncs}

\usepackage[mobile]{eccv}

\usepackage{eccvabbrv}

\usepackage{graphicx}
\usepackage{booktabs}
\usepackage{colortbl}
\usepackage{multirow}
\usepackage{makecell}

\usepackage[accsupp]{axessibility}  

\usepackage{hyperref}

\usepackage{orcidlink}
\usepackage{marvosym}

\begin{document}

\title{Enhancing Vectorized Map Perception with Historical Rasterized Maps}

\titlerunning{HRMapNet}

\author{Xiaoyu Zhang\inst{1, *}\orcidlink{0000-0002-8674-3790} \and
Guangwei Liu\inst{2, *}\orcidlink{0009-0004-1585-8090} \and
Zihao Liu\inst{3}\orcidlink{0009-0004-6680-5319} \and
Ningyi Xu\inst{3}\orcidlink{0009-0004-6809-7694} \and \\
Yunhui Liu\inst{1}\textsuperscript{,\Letter}\orcidlink{0000-0002-3625-6679} \and
Ji Zhao\inst{2,\dagger}\orcidlink{0000-0002-0150-4601}
}

\authorrunning{X.~Zhang et al.}

\institute{The Chinese University of Hong Kong \and
Huixi Technology\and
Shanghai Jiao Tong University \\
\email{zhang.xy@link.cuhk.edu.hk} \quad
\email{xuningyi@sjtu.edu.cn} \\
\email{yhliu@mae.cuhk.edu.hk} \quad
\email{zhaoji84@gmail.com}
}

\maketitle

\renewcommand{\thefootnote}{}
\footnotetext{*: Equal contribution. $\dagger$: Project lead. \Letter: Corresponding author.}

\newcommand{\name}{HRMapNet}

\input{sec/0_abstract}    
\input{sec/1_introduction}

\input{sec/2_related}
\input{sec/3_method}
\input{sec/4_experiments}
\input{sec/5_conclusion}

\section*{Acknowledgements}
This work is supported by the Shenzhen Portion of Shenzhen-Hong Kong Science and Technology Innovation Cooperation Zone 
under HZQB-KCZYB-20200089, 
the CUHK T Sone Robotics Institute, 
and the InnoHK of the Government of Hong Kong via the Hong Kong Centre for Logistics Robotics.

\input{supp}

%
%
\bibliographystyle{splncs04}
\bibliography{main}

\end{document}

%% file: sec/0_abstract.tex
\begin{abstract}
In autonomous driving, there is growing interest in end-to-end online vectorized map perception in bird's-eye-view (BEV) space, with an expectation that it could replace traditional high-cost offline high-definition (HD) maps. 
However, the accuracy and robustness of these methods can be easily compromised in challenging conditions, such as occlusion or adverse weather, when relying only on onboard sensors.
In this paper, we propose {\bf \name}, leveraging a low-cost {\bf H}istorical {\bf R}asterized {\bf Map} to enhance online vectorized map perception.
The historical rasterized map can be easily constructed from past predicted vectorized results and provides valuable complementary information.
To fully exploit a historical map, we propose two novel modules to enhance BEV features and map element queries.
For BEV features, we employ a feature aggregation module to encode features from both onboard images and the historical map. 
For map element queries, we design a query initialization module to endow queries with priors from the historical map.
The two modules contribute to leveraging map information in online perception.
Our {\name} can be integrated with most online vectorized map perception methods.
We integrate it in two state-of-the-art methods, significantly improving their performance on both the nuScenes and Argoverse~2 datasets.
The source code is released at \url{https://github.com/HXMap/HRMapNet}.
\keywords{Autonomous driving \and Bird’s-Eye-View \and Vectorized map perception \and Historical map}
\end{abstract}

%% file: sec/1_introduction.tex
\section{Introduction}
High-definition (HD) maps comprise positions and structures of vectorized map elements (e.g., lane divider, pedestrian crossing, and road boundaries), playing a vital role in the navigation of self-driving vehicles.
Traditionally, HD maps are constructed offline, utilizing SLAM-based methods~\cite{zhang2014loam, shan2018lego} and complex pipelines for annotation and vectorization.
The high cost of constructing and maintaining an HD map severely impedes the development of autonomous driving. 
Consequently, researchers are turning to online map perception using onboard sensors.

The HD map used in autonomous driving is a type of \emph{vectorized map}, which is a collection of point sets for each map element. Such vectorized representation is friendly for downstream tasks, including motion prediction~\cite{Gao_2020_CVPR} and planning.
Some existing works treat map perception as a segmentation task~~\cite{zhou2022cross, li2022hdmapnet}
and produce a \emph{rasterized map}, which is a rectangular grid of pixels recording semantic labels for each position.
However, a rasterized map lacks instance information and requires complex processing to be converted to the desired vectorized map. 

To address the limitations above, recent work MapTR~\cite{Liao2023} defines HD map perception as a point set prediction task and utilizes DETR~\cite{carion2020end} to directly predict vectorized map elements in bird's-eye-view (BEV) space.
From then, different methods~\cite{zhang2024online, yu2023scalablemap, Qiao2023, Ding_2023_ICCV} are proposed to improve online vectorized map perception.
This trend raises the expectation of potentially discarding offline HD maps in autonomous driving.
However, relying solely on onboard sensors for online map perception poses challenges. 
Some challenging conditions, including adverse weather or occlusion, can significantly impact its accuracy and robustness.

In this paper, we want to underscore the crucial role of a historical map. 
But unlike traditional high-cost HD maps, we can alleviate the requirement of the map and maintain a low-cost one, thanks to the improving performance of online map perception. 
We propose {\name}, a novel framework designed to maintain and leverage a global historical rasterized map for vectorized map perception. 
Here, we choose a rasterized map to keep historical information for the following reasons. 
1) Vectorized maps can be rasterized easily and efficiently~\cite{Lazarow_2022_CVPR, zhang2024online}.
2) It is straightforward to merge/retrieve a local rasterized map to/from a global one.
3) A rasterized map provides clear priors of where to search for desirable map elements.
4) A rasterized map takes only a small memory footprint.

\begin{figure}[t]
    \centering
    \includegraphics[width=\linewidth]{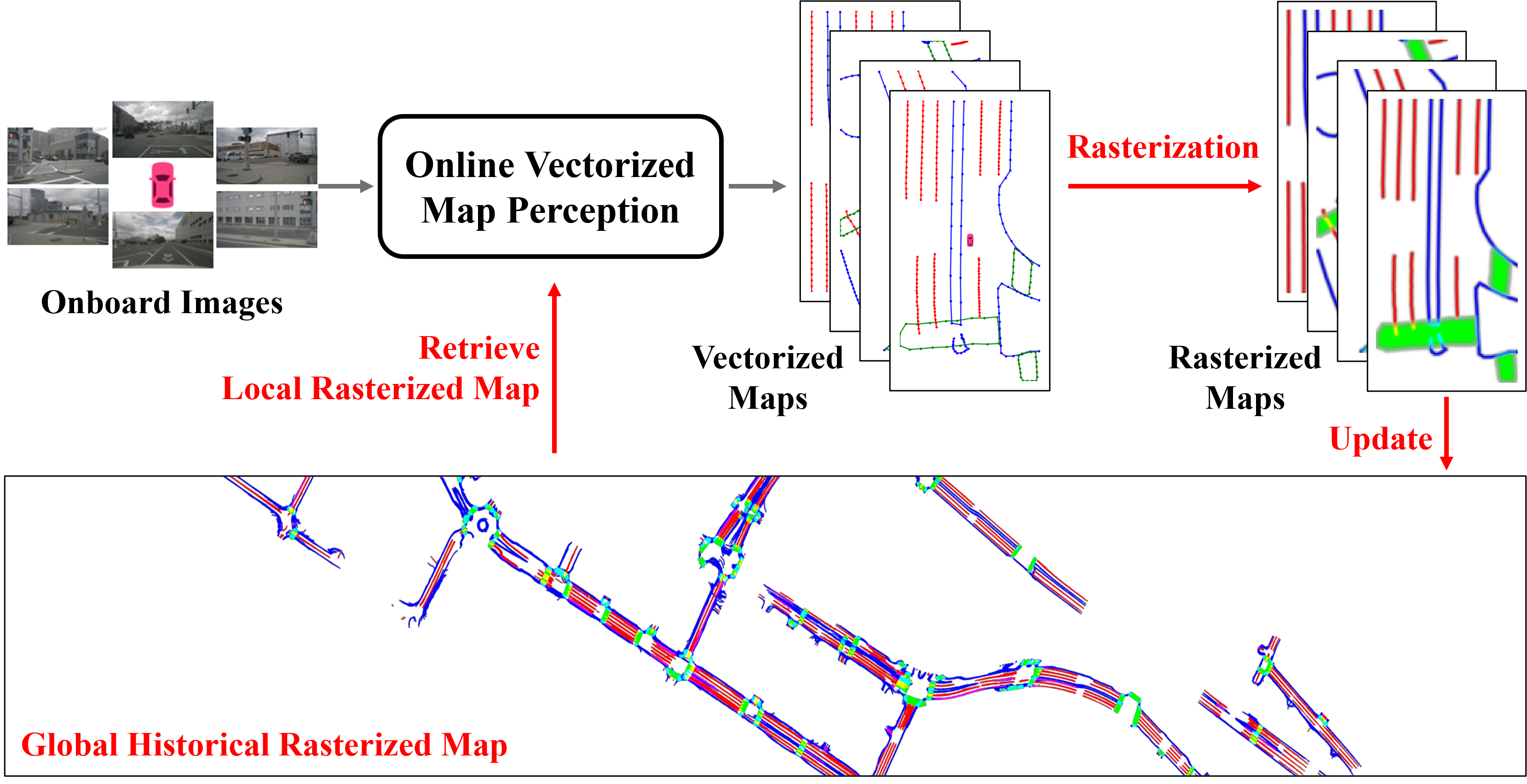}
    \caption{Pipeline of the proposed {\name}. The words in red are what we design to maintain and leverage a historical map for online perception.}
    \label{fig:pipeline}
\end{figure}

As the pipeline of {\name} in \cref{fig:pipeline}, vectorized maps from online perception are rasterized and then used to update a global historical map.
For online map perception, a local rasterized map within the current perception range is retrieved and serves as complements to onboard sensors.
The map updating and retrieving can be realized easily.
Such pipeline can be integrated with most existing state-of-the-art (SOTA) online vectorized map perception methods.

The maintained global historical rasterized map can be set from empty and updated gradually from online perception results.
When revisiting previous locations, retrieved local maps can enhance map perception by providing additional prior information.
In practice, such historical maps can even be constructed and maintained collectively by a crowd of vehicles.
Then, our method can be extended to facilitate crowdsourcing information for online map perception. 

Existing vectorized map perception methods typically encode onboard images into BEV features and use learnable queries to decode desirable map elements.
To take full advantage of historical rasterized maps within this well-established framework, we propose two novel modules to enhance both BEV features and map element queries.
Specifically, we introduce a map feature aggregation module to encode features from both images and the retrieved rasterized map, which compensates for insufficient features from onboard images alone.
Moreover, we encode the retrieved rasterized map into prior embeddings and design a query initialization module, in which base map element queries firstly interact with these map prior embeddings.
Then, the initialized queries can search for desirable map elements more efficiently.
As a result, {\name} utilizes both onboard images and a maintained historical rasterized map to achieve superior performance.
  
In summary, our main contributions are as follows:
\begin{itemize}
    \item We propose {\name}, a framework leveraging historical rasterized maps for online vectorized map perception. 
    Past predicted vectorized maps are rasterized to update a global historical rasterized map, which serves as complementary information to benefit subsequent online map perception.
    \item We design two modules to enhance both BEV features and learnable map element queries to take advantage of a historical map. 
    For BEV features, we employ a BEV feature aggregation module to encode features from both images and the retrieved rasterized map. 
    For map element queries, we design a query initialization module to search for desirable map elements efficiently.
    Both modules improve the online perception performance.
    \item We integrate {\name} with two SOTA methods (MapTRv2~\cite{Liao2023-2} and StreamMapNet~\cite{Yuan_2024_WACV}), and remarkable improvements are demonstrated on both nuScenes~\cite{Caesar_2020_CVPR} and Argoverse~2~\cite{Argoverse2} datasets under the same settings. 
    We also provide extra results to demonstrate the robustness and potential usage for practical self-driving applications. 
\end{itemize}

%% file: sec/2_related.tex
\section{Related Work}
Since map elements are commonly constructed in BEV space, online map perception benefits a lot from BEV feature learning~\cite{philion2020lift, Huang2022, li2022bevformer, Chen2022, li2023fb}, which transforms image features from surrounding cameras of a self-driving vehicle to BEV space.
For example, \cite{philion2020lift, Huang2022} lift image features to 3D space and utilize pooling to produce BEV features. 
\cite{li2022bevformer, Chen2022, li2023fb} learn BEV representations in a transformer architecture.
Map perception is generally compatible with all of these methods. 

\subsection{Map Perception with Single Frame}
At early stages, map perception mainly focuses on lane detection~\cite{Feng_2022_CVPR, Wang_2023_CVPR, Huang_2023_CVPR}, road topology reasoning~\cite{Can_2021_ICCV, Can_2022_CVPR, wang2024openlane, lanegap, li2024lanesegnet} or map segmentation~\cite{Roddick_2020_CVPR, zhou2022cross, Gosala_2023_CVPR}, which are mainly constructing rasterized maps and need post-processing to produce desired vectorized map elements for downstream applications. 
For example, predicted rasterized maps are clustered to acquire final vectorized maps in HDMapNet~\cite{li2022hdmapnet}.

VectorMapNet~\cite{pmlr-v202-liu23ax} and MapTR~\cite{Liao2023} are pioneer works to predict vectorized map elements directly.
VectorMapNet designs a map element detector and a polyline generator to produce final vectorized maps. 
MapTR proposes a unified permutation-equivalent modelling and utilizes the DETR~\cite{carion2020end} paradigm to predict vectorized map elements directly.
Following these breakthroughs, vectorized map perception becomes popular in autonomous driving research, leading to the development of many methods for improving performance.
The evolved version MapTRv2~\cite{Liao2023-2} adds decoupled self-attention in the decoder and auxiliary losses, improving the accuracy largely. 
ScalableMap~\cite{yu2023scalablemap} exploits the structural property of map elements and designs a progressive decoder for long-range perception.
MapVR~\cite{zhang2024online} introduces differentiable rasterization and rending-based loss for superior sensitivity. 
Furthermore, instead of simple point set representation, BeMapNet~\cite{Qiao2023} predicts B{\'e}zier control points and 
PivotNet~\cite{Ding_2023_ICCV} predicts pivot points instead of a fixed number of points for accurate results.

\subsection{Map Perception with Complementary Information}
The above methods predict map elements using a single frame, which limits further improvements.
Recent advancements have expanded beyond single-frame perception, incorporating complementary information.
For instance, extra standard-definition (SD) maps are explored to help HD map perception~\cite{jiang2024pmapnet} and lane-topology understanding~\cite{Luo2023}.
Satellite maps are used to augment onboard camera data for map perception in~\cite{Gao2023}.
These methods require extra data for online perception and thus increase cost in practical applications.

Temporal information is a more accessible complement for online perception. 
It has been widely used in BEV feature learning~\cite{li2022bevformer} and object detection~\cite{Wang_2023_ICCV}. 
For vectorized map perception, StreamMapNet~\cite{Yuan_2024_WACV} leverages temporal information through query propagation and BEV feature fusion.
SQD-MapNet~\cite{Wang2024} introduces stream query denoising to facilitate temporal consistency.
In these methods, short-term previous frames in temporal are utilized for perception.

One step further, if all temporal information is collected and utilized, a map can be constructed.
In~\cite{you2022hindsight}, a map of past LiDAR scans is utilized for object detection in autonomous driving.
NMP~\cite{Xiong_2023_CVPR} constructs a map of past BEV features for map segmentation.
But BEV features take substantial memory footprint, which limits its practical usage. 
Take the Boston map in nuScenes dataset~\cite{Caesar_2020_CVPR} as an example, BEV features require over 11~GB memory in NMP~\cite{Xiong_2023_CVPR}.
By contrast, we propose to maintain a low-cost historical rasterized map for vectorized map perception, which takes only 120~MB memory for the same Boston map.

%% file: sec/3_method.tex
\section{Method}
\begin{figure}[t]
    \centering
    \includegraphics[width=1\linewidth]{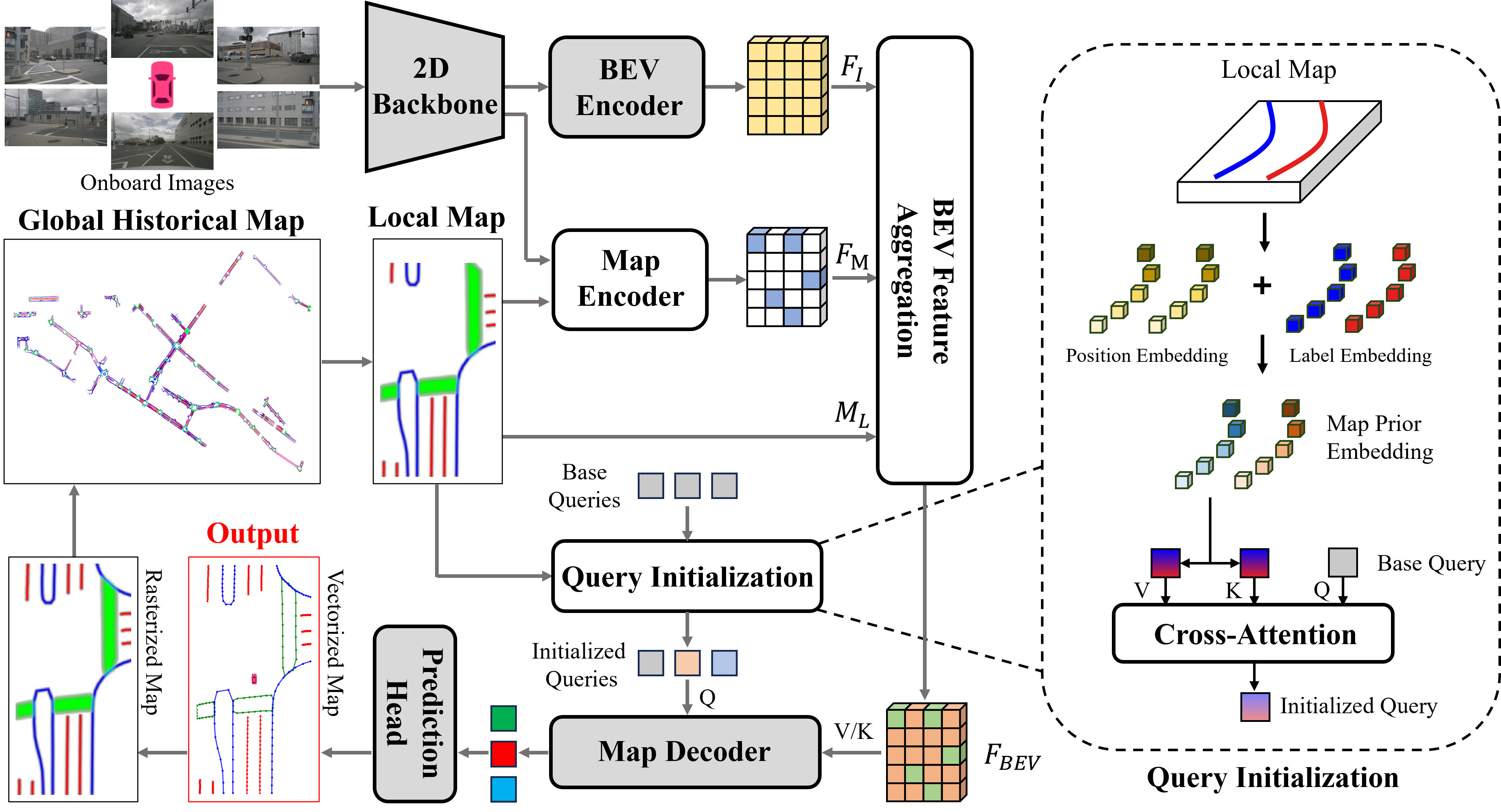}
    \caption{Architecture of the proposed {\name}. The gray blocks are kept unchanged from SOTA online vectorized map perception methods. }
    \label{fig:model}
\end{figure}

\subsection{Overview}
Our proposed {\name} is designed as a complement to SOTA online vectorized map perception methods. 
As illustrated in \cref{fig:model}, {\name} maintains a global historical rasterized map (described in \cref{sec:map}) to aid online perception.
Given surrounding images as input, 2D features are extracted from a shared backbone and transformed to BEV space. 
We introduce a map encoder and feature aggregation module (described in \cref{sec:feature}) to obtain enhanced BEV features from both onboard cameras and retrieved local maps.
Additionally, we also design a novel query initialization module (described in \cref{sec:query}), working before the original map decoder.
This module aims to endow base queries with prior information from local maps, enabling more efficient search for desirable map elements.  
Finally, vectorized map elements are predicted directly from the prediction head and can be rasterized to merge into the global map.

\subsection{Global Rasterized Map} \label{sec:map}
We firstly introduce the rasterized map used to save historical information.
As illustrated in the bottom left corner of~\cref{fig:model}, vectorized maps are rasterized to update a global map. 
Here we adopt the rasterization method used in~\cite{Lazarow_2022_CVPR}. 
Briefly, the label of each pixel in the rasterized map is determined based on its distance to vectorized elements' boundary. 
From each online prediction at $i$-th timestamp, we can obtain a semantic mask, referred as local rasterized map $M^l_i \in \{0, 1\}^{H\times{W}\times{N}}$, where $H$ and $W$ denote the spatial shape of the perception range in BEV space; $N$ is the number of map element categories and $N = 3$ as in previous methods~\cite{Liao2023, Yuan_2024_WACV}, representing lane divider, pedestrian crossing and road boundary, respectively.
Therefore, $M^l_i(p) \in \{0, 1\}^{1\times{N}} $ indicates whether and what map elements exist at position $p = (x, y)$ for each category; the value 1 indicates existence and the value 0 indicates non-existence.

Such rasterized map we utilize is analogous to occupancy grid map~\cite{1087316}, a well-established concept in robot navigation and mapping~\cite{grisetti2005improving, grisetti2007improved}.
Occupancy grid mapping~\cite{thrun2005probabilistic} is extensively studied for updating a global map from local observations.
We use a similar method to update the global rasterized map $M^g \in \mathbb{R}^{H^g\times{W^g}\times{N}}$, where $H^g$ and $W^g$ denote the spatial shape of the global map.
For map updating, the local coordinate $p$ of each pixel of $M^l_i$ is firstly transformed to the global coordinate $p^g$ based on the ego-pose $T_i= \left[R_i, t_i\right]$:
\begin{equation}
	p^g = f_{g\leftarrow l}(p; T_i) \triangleq \operatorname{round}(R_i p + t_i)
	\label{eq:g2l}
\end{equation}
where $R_i$ and $t_i$ are relative rotation and translation, and $\operatorname{round}(\cdot)$ denotes the rounding function which converts a continuous coordinate to a discrete coordinate in the rasterized map. 
Alternatively, given global coordinate $p^g$ and pose $[R_i, t_i]$, its corresponding local coordinate $p$ is
\begin{equation}
	p = f_{l\leftarrow g}(p^g; T_i) \triangleq \operatorname{round}(R_i^T (p^g - t_i))
	\label{eq:l2g}
\end{equation}

Then the global map $M^g$ can be updated based on the local map $M^l_i$ for each category. Take one category for example: 
\begin{equation}
	M^g(p^g) \leftarrow \begin{cases}
		M^g(p^g) + S^{+} & \text{if $M^l_i(p) = 1$} \\
		M^g(p^g) - S^{-} & \text{if $M^l_i(p) = 0$}
	\end{cases}
	\label{eq:update}
\end{equation}
where $p$ is determined by~\cref{eq:l2g},  $M^g$ records the status for each map element category, $S^{+}$ and $S^{-}$ are defined values to update status based on local prediction results.
This simple method integrates local results into a global map efficiently, facilitating continuous refinement and updating.
For online perception, a local rasterized map is retrieved from the global map based on the ego-pose $T_i$, and a threshold $S_{th}$ is used to determine whether map elements exist for each category:
\begin{equation}
	M^l_i(p) = \begin{cases}
		1 & \text{if $M^g(p^g) > S_{th}$} \\
		0 & \text{if $M^g(p^g) \le S_{th}$}
	\end{cases}
\end{equation}
where $p^g$ is determined by Eq. \eqref{eq:g2l}.

In our implementation, the global map $M^g$ is scaled to 8-bit unsigned int values to reduce memory consumption. 
As a result, a rasterized map consumes only a small memory footprint, about 1~MB per kilometer.

\subsection{BEV Feature Aggregation} \label{sec:feature}
Various methods~\cite{philion2020lift, Huang2022, li2022bevformer, petr2022, li2023fb} have been proposed to transform features from perspective view to BEV space, serving as a fundamental module in map perception. 
For example, MapTRv2~\cite{Liao2023-2} utilizes BEVPoolv2~\cite{Huang2022} to acquire BEV features $F_I \in \mathbb{R}^{H\times{W}\times{C}}$, where $C$ is the number of feature channels. 
We keep this module unchanged and introduce an aggregation module to enhance BEV features with prior information from local maps.

In {\name}, the retrieved local map $M^l$ serves as priors indicating where deserves more attention for map element perception. 
Therefore, inspired by FB-BEV~\cite{li2023fb}, we add additional BEV queries at locations where map elements exist in the local map (i.e., $M^l(p) \ne \mathbf{0}$).
These additional BEV queries are projected onto images to extract relevant features through spatial cross-attention~\cite{li2022bevformer}.
Then, additional BEV features are acquired where map elements exist in the local map. 
For the locations where no map elements exist (i.e., $M^l(p) = \mathbf{0}$), corresponding additional BEV features are set as zeros.
These additional BEV features are formulated as $F_M \in \mathbb{R}^{H\times{W}\times{C}}$. 

In the feature aggregation module, $F_I$, $F_M$ and $M^l$ are fused together:
\begin{equation}
    F_{BEV} = \operatorname{Conv}(\operatorname{Concat}(F_I + F_M, M^l))
\end{equation}
Here, $F_M$ is added to $F_I$ to compensate for deficiencies. 
Additionally, $M^l$ can be regarded as special BEV features with clear semantic information.
Thus we concatenate it with BEV features and use convolution to acquire the enhanced BEV features $F_{BEV} \in \mathbb{R}^{H\times{W}\times{C}}$ for further processing.

\subsection{Query Initialization} \label{sec:query}
In addition to fusing the retrieved rasterized map into BEV features, a novel query initialization module is designed to facilitate efficient search for desirable map elements.
Within a DETR~\cite{carion2020end} paradigm, a set of learnable queries will interact with extracted features to search for desirable elements. 
Without prior information, queries would search from random and prediction results are refined gradually through several decoder layers.

In {\name}, the retrieved rasterized map provides prior information about where map elements may exist and thus can facilitate map element queries search for desirable elements efficiently. 
As illustrated on the right of \cref{fig:model}, the proposed query initialization works before the original map decoder.
Base queries will firstly interact with prior features embedded from the local map.

In detail, for a valid location where map elements exist in the retrieved local map $M^l$, its 
position $p$ is related to a learnable position embedding $PE(p) \in \mathbb{R}^{1\times{C}}$; and the semantic vector $M^l(p)$ is projected to a label embedding $LE(p) \in \mathbb{R}^{1\times{C}}$ using a linear projection. 
Then, we acquire a map prior embedding for each valid position:
\begin{equation}
    ME(p) = PE(p) + LE(p)
\end{equation}
Map prior embedding $ME(p) \in \mathbb{R}^{1\times{C}}$ encodes where map elements may exist based on the retrieved local map.
To fuse priors, base queries interact with a set of map prior embeddings through cross-attention~\cite{vaswani2017attention}.
Then, initialized queries search desirable elements in BEV features through original decoder layers. Moreover, to improve efficiency and save memory consumption, the retrieved local rasterized map $M^l$ is downsampled before extracting map prior embeddings.

\subsection{Implementation Details}
{\bf Training.} The prediction head and training loss remain identical to SOTA vectorized map perception methods.
Taking MapTRv2~\cite{Liao2023-2} as an example, the prediction head predicts a class score and sequential 2D point positions for each element.
Classification loss, point-to-point loss and edge direction loss are used for training; one-to-many loss~\cite{Jia_2023_CVPR}, dense prediction loss and depth loss are used as auxiliary supervision.
In each training epoch, the global map is set from empty and updated gradually from prediction results. 

\vspace{0.5em}
\noindent{\bf Testing.} To ensure fair comparison, the global map is also  initialized as {\bf empty} by default and updated from prediction results. 
Since testing frames are typically evaluated in chronological order, most testing frames can still benefit from the updated global map from their preceding frames.

%% file: sec/4_experiments.tex
\section{Experiments} \label{sec:exper}
To demonstrate the effectiveness, we integrate {\name} with two SOTA online vectorized map perception methods, MapTRv2~\cite{Liao2023-2} and StreamMapNet~\cite{Yuan_2024_WACV}. 

\subsection{Experimental Setup}
\noindent {\bf Datasets.} 
We evaluate {\name} on two commonly used self-driving datasets, nuScenes~\cite{Caesar_2020_CVPR} and Argoverse~2~\cite{Argoverse2}.
The nuScenes dataset provides 6 surrounding images captured across 1000 scenes in 4 locations, while Argoverse~2 provides 7 surrounding images captured across 1000 scenes in 6 cities.
The two datasets comprise multiple traversals in both training and validation sets, providing diverse and comprehensive data for evaluation.

\vspace{0.5em}
\noindent {\bf Metric.}
Following MapTRv2 and StreamMapNet, three map element categories (i.e., lane divider, pedestrian crossing and road boundary) are predicted. 
Chamfer distance is used to determine whether the prediction matches with the ground truth under three thresholds (0.5 m, 1.0 m, and 1.5 m). 
The mean average precision (mAP) is calculated for the three categories.

\vspace{0.5em}
\noindent {\bf Details.}
We keep all training and validation details the same as MapTRv2 and StreamMapNet. 
We use MapTRv2 as an example to elucidate the subsequent settings, more details can be found in their original papers.

Each map element is modelled as 20 sequential points. 
The perception range is set as [-30m, 30m] from rear to front and [-15m, 15m] from left to right, and the resolution of BEV features is 0.3m$\times$0.3m.
The shape of a local rasterized map is set the same as BEV features; the resolution of both local and global rasterized maps is also 0.3m$\times$0.3m.
We set $S^{+}$ as 30 and $S^{-}$ as 1 to update the global map.

Our model is trained on 8 NVIDIA A100 GPUs with a batch size of $8\times4$, the learning rate is set to $6\times10^{-4}$. 
ResNet50~\cite{he2016deep} is used as the backbone to extract image features. 

\subsection{Comparison with SOTA Methods}
\newcommand{\upperf}[1]{$^{\textcolor{red}{(\uparrow #1)}}$}
\definecolor{myblue}{rgb}{.78, .96, 1.0}
\begin{table}[t]
  \centering
  \caption{Comparison on nuScenes~\cite{Caesar_2020_CVPR}. 
  In each block, the first row is the method as baseline and improved methods are labelled by ``\#''. 
  In the ``Modality'' column, - means using only a single frame; ``SDMap'' means adding an extra SDMap as input; ``Temporal'' means using temporal information; ``HRMap'' denotes using a historical rasterized map.
  The results of MapTR and P-MapNet are taken from P-MapNet; the results of StreamMapNet and SQD-MapNet are taken from SQD-MapNet, which are also consistent with the results reproduced by ourselves. Other results are taken from their papers.
  FPS is measured on a single NVIDIA A100 GPU with batch size of 1.
  The improvements introduced by our method are labelled in red.}
    \begin{tabular}{lcc|cccc|c}
    \toprule
    \makebox[2cm][c]{Method} & \makebox[1.8cm][c]{Modality} & \makebox[1.0cm][c]{Epoch} & \makebox[1.1cm][c]{AP$_{ped}$} &
    \makebox[1.1cm][c]{AP$_{div}$} & \makebox[1.1cm][c]{AP$_{bou}$ } & \makebox[1.1cm][c]{mAP} & \makebox[1cm][c]{FPS} \\
    \midrule
    VectorMapNet~\cite{pmlr-v202-liu23ax} & -     & 110    & 36.1  & 47.3  & 39.3  & 40.9  & - \\
    PivotNet~\cite{Ding_2023_ICCV} & -     & 24    & 56.2  & 56.5  & 60.1  & 57.6  & - \\
    BeMapNet~\cite{Qiao2023} & -     & 30    & 57.7  & 62.3  & 59.4  & 59.8  & - \\
    \midrule
    MapTR~\cite{Liao2023} & -     & 24    & 41.2  & 49.5  & 51.1  & 47.3  & - \\
    \# P-MapNet~\cite{jiang2024pmapnet} & SDMap & 24    & 43.7  & 50.9  & 53.5  & 49.4  & - \\
    \midrule
    StreamMapNet~\cite{Yuan_2024_WACV} & -     & 24    & 60.4  & 61.9  & 58.9  & 60.4  & 22.5 \\
    \# SQD-MapNet~\cite{Wang2024}  & Temporal  & 24    & 63.0  & 65.5  & 63.3  & 63.9  & - \\
    \rowcolor{myblue} \# {\name} & HRMap & 24    & \bf 63.8  &\bf 69.5  &\bf 65.5  & \bf 66.3\upperf{5.9} & 21.1 \\
    \midrule
    MapTRv2~\cite{Liao2023-2} & -     & 24    & 59.8  & 62.4  & 62.4  & 61.5  & 19.6 \\
    MapTRv2~\cite{Liao2023-2} & -     & 110   & 68.1  & 68.3  & 69.7  & 68.7  & 19.6 \\
    \rowcolor{myblue} \# {\name} & HRMap & 24    &\bf 65.8  &\bf 67.4  &\bf 68.5  & \bf 67.2\upperf{5.7} & 17.0 \\
    \rowcolor{myblue} \# {\name} & HRMap & 110   &\bf 72.0  &\bf 72.9  &\bf 75.8  & \bf 73.6\upperf{4.9} & 17.0 \\
    \bottomrule
    \end{tabular}%
  \label{tab:nuscenes}%
\end{table}%
\noindent {\bf Comparison on nuScenes.}
As illustrated in \cref{tab:nuscenes}, we compare {\name} with SOTA vectorized map perception methods.
{\name} largely outperforms the methods using only single frame images (i.e.,  VectorMapNet, PivotNet, BeMapNet, MapTR, StreamMapNet and MapTRv2).
Specifically, when trained with 24 epochs, {\name} boosts the performance of StreamMapNet and MapTRv2 by $+5.9$ mAP and $+5.7$ mAP, respectively.
When trained with 110 epochs, {\name} still outperforms MapTRv2 by $+4.9$ mAP under the same setting.

In comparison to methods introducing complementary information, {\name} also stands out for its comprehensive utilization of all historical information.
Besides onboard images, P-MapNet adds extra SDMap from OpenStreetMap~\cite{haklay2008openstreetmap}, but the improvement is lower than ours.
SQD-MapNet is a concurrent work which leverages stream query denoising strategy to benefit from the results of previous frames.
Since all predicted results are preserved in a global rasterized map, {\name} leverages not only temporal information but all past results for online perception, and thus achieves superior performance.

Moreover, {\name} maintains high efficiency in terms of inference speed when integrated with StreamMapNet and MapTRv2, running at 21.1 FPS and 17.0 FPS, respectively. 
This ensures {\name} not only enhances performance but also remains practical for real-time applications in autonomous driving.

\vspace{0.5em}
\noindent {\bf Comparison on Argoverse~2.}
The results on Argoverse~2 dataset, as presented in \cref{tab:av2}, further demonstrate the effectiveness of {\name}.
{\name} achieves significant enhancements across both StreamMapNet and MapTRv2, with an increase of +2.8 mAP for StreamMapNet, surpassing the performance of SQD-MapNet which leverages temporal information. 
Similarly, when compared to MapTRv2, {\name} exhibits superior performance with a gain of +4.0 mAP. 
These results underscore the effectiveness and versatility of our method across different methodologies and datasets.

\begin{table}[t]
  \centering
  \caption{Comparison on Argoverse~2~\cite{Argoverse2}. $^\dagger$: The results are re-evaluated using the official codes under the same setting with StreamMapNet. }
    \begin{tabular}{lcc|cccc|c}
    \toprule
    \makebox[2cm][c]{Method} & \makebox[1.8cm][c]{Modality} & \makebox[1cm][c]{Epoch} & \makebox[1.1cm][c]{AP$_{ped}$} &
    \makebox[1.1cm][c]{AP$_{div}$} & \makebox[1.1cm][c]{AP$_{bou}$ } & \makebox[1.1cm][c]{mAP} & \makebox[1.cm][c]{FPS} \\
    \midrule
    StreamMapNet~\cite{Yuan_2024_WACV} & -     & 30    & 62.0  & 59.5  & 63.0  & 61.5  & 20.2 \\
    \# SQD-MapNet~\cite{Wang2024}  & Temporal  & 30    &\bf 64.9  & 60.2  & 64.9  & 63.3  & - \\
    \rowcolor{myblue} \# {\name} & HRMap & 30    & 63.8  &\bf 62.5  &\bf 66.6  & \bf 64.3\upperf{2.8} & 19.5 \\
    \midrule
    MapTRv2$^\dagger$~\cite{Liao2023-2} & -     & 30    & 60.0  & 68.7  & 64.2  & 64.3  & 18.1 \\
    \rowcolor{myblue} \# {\name} & HRMap & 30   &\bf 65.1  & \bf 71.4  &\bf 68.6  & \bf 68.3 \upperf{4.0} & 16.2 \\
    \bottomrule
    \end{tabular}%
  \label{tab:av2}%
\end{table}%

\begin{table}[t]
  \centering
  \caption{Comparisons on new split data sets proposed in StreamMapNet.}
    \begin{tabular}{c|lc|cccc}
    \toprule
    \makebox[2.1cm][c]{Dataset} & \makebox[2cm][c]{Method} & \makebox[1.8cm][c]{Modality} & \makebox[1.1cm][c]{AP$_{ped}$} &
    \makebox[1.1cm][c]{AP$_{div}$} & \makebox[1.1cm][c]{AP$_{bou}$ } & \makebox[1.1cm][c]{mAP} \\
    \midrule
    \multirow{2}{*}{nuScenes} & StreamMapNet~\cite{Yuan_2024_WACV} & Temporal   & 29.6  & 30.1  & 41.9  & 33.9\\
     & \cellcolor{myblue}\# {\name} &\cellcolor{myblue} HRMap  &\cellcolor{myblue} \bf 36.9  &\cellcolor{myblue}\bf 30.3  &\cellcolor{myblue}\bf 44.0  & \cellcolor{myblue}\bf 37.1\upperf{3.2} \\
    \midrule
    \multirow{2}{*}{Argoverse~2} & StreamMapNet~\cite{Yuan_2024_WACV} & Temporal  & 56.9  & 55.9  & 61.4  & 58.1\\
     &\cellcolor{myblue}\# {\name} &\cellcolor{myblue} HRMap  &\cellcolor{myblue} \bf 60.1  & \cellcolor{myblue}\bf 58.3  &
     \cellcolor{myblue}\bf 66.0  &\cellcolor{myblue} \bf 61.5\upperf{3.4} \\
    \bottomrule
    \end{tabular}%
  \label{tab:new}%
\end{table}%

\vspace{0.5em}
\noindent {\bf Comparison on new split sets.} \label{sec:new}
The above experiments are conducted on the commonly used original dataset split, in which overlap of locations exist between training and validation sets. 
StreamMapNet proposes new splitting methods for nuScenes and Argoverse~2, in which training and validation data are separated in locations.
We also provide results on these new split sets in \cref{tab:new}. 
On the new split data, StreamMapNet utilizes query propagation and BEV fusion to integrate temporal information.
We do not use these two temporal fusion modules and only integrate utilizing a global rasterized map. 
{\name} still improves the base StreamMapNet by over +3.0 mAP on these two datasets, which reinforces the value of integrating a global rasterized map.

\subsection{Ablation Study} \label{sec:ablation}
In this subsection, we provide some ablation studies of our method, which are conducted on nuScenes with 24 epochs and use MapTRv2~\cite{Liao2023-2} as the baseline. 

\vspace{0.5em}
\noindent {\bf Feature aggregation and query initialization.}
For online perception, our method leverages global map information in both BEV features and queries. 
We provide an ablation study about these two modules in \cref{tab:ablation1}.
From MapTRv2, integrating global map information into BEV features brings an improvement of +3.1~mAP.
Introducing query initialization further improves the performance by +2.6~mAP.
Both components have significant positive effect for integrating global map information into online perception.

To further demonstrate that the proposed query initialization helps search for map elements more efficiently, an ablation study of decreasing decoder layers is provided in the supplementary material.

\begin{table}[t]
  \centering
  \caption{Ablations on each component of HRMap. The improvement introduced by each component is labelled in red. }
    \begin{tabular}{l|cccc}
    \toprule
    \makebox[3.5cm][c]{Method} & \makebox[1.2cm][c]{AP$_{ped}$} & \makebox[1.2cm][c]{AP$_{div}$} & \makebox[1.2cm][c]{AP$_{bou}$ } & \makebox[1.5cm][c]{mAP} \\
    \midrule
     MapTRv2~\cite{Liao2023-2}    & 59.8  & 62.4  & 62.4  & 61.5\\
     + Feature Aggregation   & 62.6 & 64.5 & 66.8 & 64.6\upperf{3.1} \\
     + Query Initialization  & 65.8  & 67.4  & 68.5  &  67.2\upperf{2.6}\\
    \bottomrule
    \end{tabular}%
  \label{tab:ablation1}%
\end{table}%
\begin{table}[t]
  \centering
  \caption{Ablations on map resolution in query initialization. The default setting is highlighted in blue. Mem denotes the maximum GPU memory consumed in training.}
    \begin{tabular}{lc|cccccc}
    \toprule
    \makebox[1.6cm][c]{Method} & \makebox[1.6cm][l]{Resolution} & \makebox[1.2cm][c]{AP$_{ped}$} & \makebox[1.2cm][c]{AP$_{div}$} & \makebox[1.2cm][c]{AP$_{bou}$ } & \makebox[1.2cm][c]{mAP} 
    & \makebox[1.6cm][c]{Mem. (GB)} & \makebox[1cm][c]{FPS} \\
    \midrule
     MapTRv2 & - & 59.8  & 62.4  & 62.4  & 61.5 & 22.25 & 19.6\\
     \multirow{5}{*}{\#{\name}} & 0.3 m & 63.9 & 66.5 & 68.6 & 66.3 & 64.94 & 16.6\\
     & \cellcolor{myblue}{0.6 m} & \bf \cellcolor{myblue}65.8 &\bf \cellcolor{myblue}67.4 &\cellcolor{myblue}68.5 & \bf \cellcolor{myblue}67.2 & \cellcolor{myblue}31.41 & \cellcolor{myblue}17.0\\
    & 0.9 m & 64.1 & 65.3 &\bf 69.6 & 66.4 & 25.17 & 17.1\\
    & 1.2 m & 63.2 & 67.2 & 69.4 & 66.6 & 25.22 & 17.1\\
    & 1.5 m & 64.3 & 65.6 & 68.2 & 66.0 & 25.94 & 17.2\\
    \bottomrule
    \end{tabular}%
  \label{tab:ablation2}%
\end{table}%

\vspace{0.5em}
\noindent {\bf Map resolution in query initialization.}
In query initialization, all valid positions where map elements exist are embedded as map priors to endow information to base queries. 
However, sometimes too many prior embeddings are extracted, consuming large amounts of memory.
To address this, local rasterized maps are downsampled to a coarser resolution before extracting map prior embeddings.
In~\cref{tab:ablation2}, we provide an ablation study of the resolution used to encode map prior embeddings.
For the resolution of 0.3 m, downsampling is not used, resulting in a large number of embeddings and significant GPU memory consumption during training.
As the resolution decreases, memory usage also decreases rapidly.
The impact on inference speed is minimal. 
We set the resolution to 0.6 m as the default setting, because the memory consumption is acceptable and it gets the best performance.
The results also indicate {\name} consumes about 9~GB extra GPU memory in training, compared to MapTRv2.

\subsection{Extra Results for Practical Usage}
\begin{table}[t]
  \centering
  \caption{Testing mAP with different localization errors. The same model is tested with varying levels of noise.
  The random noise is subject to a normal distribution. $\sigma_t$ and $\sigma_r$  are applied standard deviations for translation and rotation respectively.
  }
    \begin{tabular}{cc|cccc}
    \toprule
    & & \multicolumn{4}{c}{$\sigma_t$ (m)} \\
    \makebox[1cm][c]{} & \makebox[1.2cm][c]{} & \makebox[1.2cm][c]{0} & \makebox[1.2cm][c]{0.05} & \makebox[1.2cm][c]{0.1} & \makebox[1.2cm][c]{0.2} \\
    \midrule
    \multirow{4}{*}{\makecell{$\sigma_r$ \\ (rad)}} 
    & 0 & \bf 67.2 & 67.1 & 66.9 & 66.2 \\
    & 0.005 & 66.7 & 66.7 & 66.6 & 66.0 \\
    & 0.01 & 66.2 & 66.0 & 66.0 & 65.4\\
    & 0.02 & 64.6 & 64.5 & 64.2 & 63.8\\
    \bottomrule
    \end{tabular}%
  \label{tab:localization}%
\end{table}%
\noindent {\bf Robustness to localization error.}
As described in \cref{sec:map}, the global map is updated from local predicted rasterized maps based on ego-pose. 
In autonomous driving, ego-pose can be localized with high accuracy using GNSS modules or SLAM-based methods~\cite{zhang2014loam, shan2018lego}. 
To assess the robustness of {\name} to localization errors, we conduct additional experiments on the nuScenes dataset, as presented in \cref{tab:localization}. 
The model is based on MapTRv2 and trained with 24 epochs; all results are from the same model with varying levels of localization errors.
We add random noises to both translation and rotation of ego-pose, and thus both map updating and retrieving would be affected.

The results clearly demonstrate the robustness of {\name} to localization errors, particularly in terms of translation. 
One contributing factor is the relatively small map resolution (0.3m) utilized in our method. 
Despite varying levels of localization errors, {\name} consistently achieves comparable results, experiencing only a 1 mAP drop in most cases.
Even in the worst case with the largest noise, a historical map still brings benefits (63.8 mAP) compared to the baseline, MapTRv2 (61.5 mAP).

Considering localization with 0.1 m error for translation and 0.01 rad error for rotation is a common requirement in autonomous driving~\cite{reid2019localization}, these extra results indicate the effectiveness of {\name} in practical usage.

\vspace{0.5em}
\noindent {\bf Different initial maps.}
In the above experiments, {\name} is tested with an empty initial global map for a more fair comparison.
The global map is updated gradually from perception results and benefits later prediction.
For many frames, the online perception actually only benefits from short-term previous frames in temporal, which weakens the power of using a global historical map.

\begin{table}[t]
  \centering
  \caption{Ablations on initial maps. 
  The model used here is the same with only different initial maps.
  ``Empty'' is the default setting. ``Validation Map'' denotes using validation data to construct a global map for the first running. ``Training Map'' denotes the global map constructed during training.}
    \begin{tabular}{c|cccc}
    \toprule
    \makebox[3.5cm][c]{Initial Map} & \makebox[1.2cm][c]{AP$_{ped}$} & \makebox[1.2cm][c]{AP$_{div}$} & \makebox[1.2cm][c]{AP$_{bou}$ } & \makebox[1.5cm][c]{mAP} \\
    \midrule
     \rowcolor{myblue} Empty    & 65.8  & 67.4  & 68.5  & 67.2\\
     Validation Map & 72.0  & 71.9  & 73.9  &  72.6\\
     Training Map   & 81.8 & 85.9 & 83.4 & 83.7 \\
    \bottomrule
    \end{tabular}%
  \label{tab:map}%
\end{table}%

Here, we provide extra results with pre-built initial maps in \cref{tab:map}.
The model used here is the same as in~\cref{tab:nuscenes}, integrated with MapTRv2 and trained 24 epochs.
Note that the model is {\bf not re-trained or finetuned}, and we only test it with different initial maps. 
Here are two kinds of maps can be provided. 

For the ``validation map'', the same model is tested with validation data twice.
The first time is running with an empty initial map.
The map is updated gradually as validation data comes in and the final global map is saved.
This global map is loaded for the second validation.
There is actually no extra data input, the model constructs a global map by itself and use it again for validation.
With the help of this more complete map, the performance of the same model is further enhanced by +5.4~mAP.

Besides, the global map constructed during training is saved and loaded again for validation.
As stated in \cref{sec:new}, there are overlaps in location between training and validation data.
Thus this training map can also benefit online perception for validation.
Because this training map is more accurate, the performance is improved largely by +16.5~mAP.

We provide these extra results to show the potential of {\name} for practical usage in autonomous driving, including crowdsourcing online map perception.
Provided with an easily maintained global map, which may even be constructed by other vehicles, the performance of online vectorized map perception can be improved largely. 

\subsection{Qualitative Results.}
\begin{figure}[t]
    \centering
    \includegraphics[width=1\linewidth]{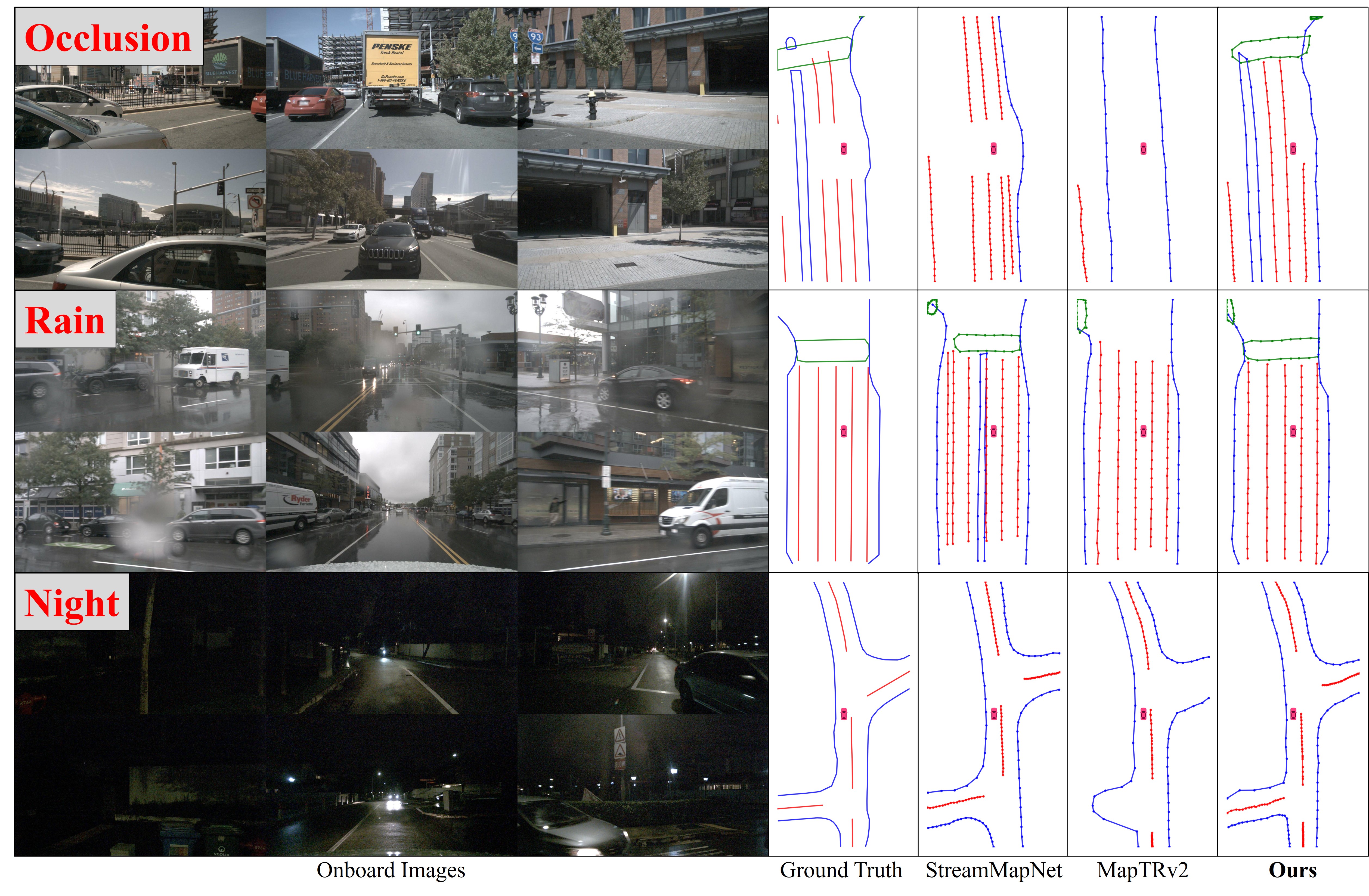}
    \caption{Visualized results in three challenging scenarios: severe occlusion, rainy day and poor lighting at night. All three methods are trained with 24 epochs. Ours is based on MapTRv2. 
    Lane divider, pedestrian crossing and road boundary are illustrated in red, green and blue respectively.}
    \label{fig:examples}
\end{figure}

In \cref{fig:examples}, we show some qualitative comparisons in three challenging scenarios: severe occlusion, rainy day and poor lighting at night. 
The online map perception relying only on onboard sensors can be easily affected by these inevitable factors.
Leveraging a historical rasterized map, {\name} helps to improve online map perception ability to handle such challenges.
More visualized results and analysis are included in the supplementary material.

%% file: sec/5_conclusion.tex
\section{Discussion and Conclusion}
In this paper, we propose to leverage historical information by maintaining a global rasterized map for improved online vectorized map perception. 
The global rasterized map can be constructed and maintained easily from past prediction results. 
We utilize such historical rasterized maps as complementary information for both BEV feature aggregation and query initialization.
The proposed framework is compatible with most existing online vectorized map perception methods.
It is demonstrated our proposed {\name} can boost two SOTA online vectorized map perception methods by a large margin.
We expect {\name} can be a basis for crowdsourcing map perception: an accurate global rasterized map is maintained by a crowd of self-driving vehicles and serves as priors for accurate online vectorized map perception for each vehicle.

\vspace{0.5em}
\noindent {\bf Limitations.}
Our proposed {\name} mainly focuses on how to leverage a historical rasterized map for online vectorized map perception.
We do not design elaborate map maintaining methods and only use a simple yet effective one from robotic occupancy grid mapping to merge local predictions to a global map.
In practice, more intelligent methods could be explored and utilized, such as \cite{9197261} for collaborative semantic mapping; \cite{9561375} utilizing recurrent neural networks; \cite{xie2023mv} produces consistent rasterized maps from multiple predictions.

%% file: supp.tex
\definecolor{myblue}{rgb}{.78, .96, 1.0}

\appendix
\section*{Appendix}
\section{Performance under Original MapTRv2 Setting}
Here we provide the validation results on Argoverse~2 dataset in \cref{tab:av2_2}, under the original MapTRv2 setting. Here all training data at 10~Hz is used for training, and 2.5~Hz validation data is extracted for validation. Our {\name} still outperforms the baseline, MapTRv2. 

\begin{table}[h]
  \centering
  \caption{Comparison on Argoverse~2~\cite{Argoverse2} under the original MapTRv2 setting. }
    \begin{tabular}{lc|cccc|c}
    \toprule
    \makebox[2cm][c]{Method}  & \makebox[1.5cm][c]{Epoch} & \makebox[1.5cm][c]{AP$_{ped}$} &
    \makebox[1.5cm][c]{AP$_{div}$} & \makebox[1.5cm][c]{AP$_{bou}$ } & \makebox[1.5cm][c]{mAP} & \makebox[1.5cm][c]{FPS} \\
    \midrule
    MapTRv2~\cite{Liao2023-2}  & 6    & 62.9  &\bf 72.1  & 67.1  & 67.4  & 18.1 \\
    \rowcolor{myblue} \# {\name} & 6   &\bf 64.1  & 71.5  &\bf 69.7  & \bf 68.5\upperf{1.1} & 16.2 \\
    \bottomrule
    \end{tabular}%
  \label{tab:av2_2}%
\end{table}%

\section{Performance under Challenging Conditions}
To further demonstrate the improvements of utilizing a HRMap, we summarize the performance under different conditions in \cref{tab:challenge}. It is clear the improvement is significant especially for these challenging conditions.
\begin{table}[h]
  \centering
  \caption{Performance under different conditions on nuScenes~\cite{Caesar_2020_CVPR}, trained with 24 epochs.}
    \begin{tabular}{lc|ccc}
    \toprule
    \makebox[2.0cm][c]{Method} & \makebox[3cm][c]{Initial Map} & \makebox[2.0cm][c]{Night} & \makebox[2.0cm][c]{Rainy} & \makebox[2.0cm][c]{Normal}\\
    \midrule
     MapTRv2 & -  & 39.6 & 50.8 & 64.6\\
     \midrule
     \multirow{2}{*}{\# {\name}} 
     & Empty & 42.6\upperf{3.0} & 62.6\upperf{11.8} & 69.4\upperf{5.2}\\
     & Training Map & 74.8\upperf{35.2} & 72.9\upperf{22.1} & 85.9\upperf{21.3}\\
    \bottomrule
    \end{tabular}%
  \label{tab:challenge}%
\end{table}%

\section{Performance for Long Range}
We test {\name} (StreamMapNet~\cite{Yuan_2024_WACV} as baseline) by just increasing perception range. 
As in \cref{tab:long}, with online constructed map, {\name} also boosts the baseline a lot. 
For such far ranges, we suggest loading a pre-built map like in P-MapNet~\cite{jiang2024pmapnet} for practical usage.
We further test {\name} with a pre-built map from training data, and mAP is improved to 57.3 and 40.2, respectively.
Besides, we believe {\name} can achieve better results with more careful settings, such as suitable map resolution, query number.
\begin{table}[h]
  \centering
  \caption{Evaluation under long range on nuScenes~\cite{Caesar_2020_CVPR}. The results of MapTR and P-MapNet are from the paper of P-MapNet. StreamMapNet is trained and evaluated using the official codes. Our {\name} is based on StreamMapNet. $^{\dagger}$: Performance with a loaded map built from training data. All models are trained with 24 epochs.}
    \begin{tabular}{cl|ccc|c}
    \toprule
    \makebox[2.0cm][c]{Range} & \makebox[3cm][c]{Method} & \makebox[1.2cm][c]{AP$_{ped}$} & \makebox[1.2cm][c]{AP$_{div}$} & \makebox[1.2cm][c]{AP$_{bou}$ } & \makebox[1.5cm][c]{mAP} \\
    \midrule
     \multirow{5}{*}{120$\times$60m} 
     & MapTR~\cite{Liao2023} & 18.9 & 26.0 & 15.7 & 20.2 \\
     &\# P-MapNet~\cite{jiang2024pmapnet} & 22.0 & 27.2 & 19.5 & 22.9 \\
     &StreamMapNet~\cite{Yuan_2024_WACV} & 37.2 & 42.3 & 30.2 & 36.6 \\
     &\cellcolor{myblue}\# {\name} &\cellcolor{myblue}\bf 43.1  &\cellcolor{myblue}\bf 47.7 &\cellcolor{myblue}\bf 34.9  & \cellcolor{myblue}\bf41.9\upperf{5.3} \\
     &\cellcolor{myblue}\# {\name}$^{\dagger}$ &\cellcolor{myblue} 65.0  &\cellcolor{myblue} 61.9 &\cellcolor{myblue} 44.9  & \cellcolor{myblue} 57.3 \\
         \midrule
     \multirow{5}{*}{240$\times$120m} 
     & MapTR~\cite{Liao2023} & 7.2 & 12.7 & 4.2 & 8.0 \\
     &\# P-MapNet~\cite{jiang2024pmapnet} & 16.3 & 22.7 & 10.5 & 16.5 \\
     &StreamMapNet~\cite{Yuan_2024_WACV} & 22.4 & 31.3 & 16.0 & 23.3 \\
     &\cellcolor{myblue}\# {\name} &\cellcolor{myblue}\bf 29.4  &\cellcolor{myblue}\bf 34.2 &\cellcolor{myblue}\bf 19.7  & \cellcolor{myblue}\bf27.8\upperf{4.5} \\
     &\cellcolor{myblue}\# {\name}$^{\dagger}$ &\cellcolor{myblue}51.0  &\cellcolor{myblue}46.2 &\cellcolor{myblue}23.5  & \cellcolor{myblue}40.2 \\
    \bottomrule
    \end{tabular}%
  \label{tab:long}%
\end{table}%

\section{Ablation on Decoder Layers}
Our method keeps the map decoder module the same as the baseline methods~\cite{Liao2023-2, Yuan_2024_WACV}, and 6 decoder layers are used in these methods.
With the help of retrieved rasterized map as priors, the proposed query initialization module helps map element queries search for desirable map elements more efficiently. 
Here, we provide an ablation study of the number of decoder layers in~\cref{tab:layer} to further demonstrate the searching efficiency improved by this module.
The results of not using query initialization are also listed for comparison.

Increasing decoder layers commonly brings higher accuracy. 
The comparison indicates using only 4 decoder layers with query initialization have already achieved good results, better than the method using 6 decoder layers without query initialization.
It is because query initialization endows map element queries with priors.

\definecolor{myblue}{rgb}{.78, .96, 1.0}
\begin{table}[h]
  \centering
  \caption{Ablation of the number of decoder layers on nuScenes~\cite{Caesar_2020_CVPR}, trained with 24 epochs. The results of the method without query initialization are listed for comparison. The results better than the method without query initialization are highlighted in blue.}
    \begin{tabular}{l|c|cccc}
    \toprule
    \makebox[4.0cm][c]{Method} & \makebox[1.2cm][c]{Layer} & \makebox[1.2cm][c]{AP$_{ped}$} & \makebox[1.2cm][c]{AP$_{div}$} & \makebox[1.2cm][c]{AP$_{bou}$ } & \makebox[1.2cm][c]{mAP} \\
    \midrule
     w/o Query Initialization & 6   & 62.6 & 64.5 & 66.8 & 64.6\\
     \midrule
     \multirow{4}{*}{w/ Query Initialization} 
       & 3 & 60.7 & 61.9 & 65.4 & 62.6 \\
       &\cellcolor{myblue}4 &\cellcolor{myblue}63.6  &\cellcolor{myblue}64.7  &\cellcolor{myblue}68.0  & \cellcolor{myblue}65.4\\
       &\cellcolor{myblue}5 &\cellcolor{myblue}64.5  &\cellcolor{myblue}66.5 &\cellcolor{myblue}68.6  & \cellcolor{myblue}66.6\\
       &\cellcolor{myblue}6 &\cellcolor{myblue}65.8  &\cellcolor{myblue}67.4  &\cellcolor{myblue}68.5  & \cellcolor{myblue}67.2\\
    \bottomrule
    \end{tabular}%
  \label{tab:layer}%
\end{table}%

\section{Discussion about Localization Error}
By default, we train and test HRMapNet with the groundtruth ego-pose, which is a common practice in temporal-based map perception and object detection methods, such as StreamMapNet~\cite{Yuan_2024_WACV} and BEVFormer~\cite{li2022bevformer}.
In the main body of the paper, it is demonstrated that HRMapNet has certain robustness to pose error even without specific design.
The noise level is set according to common requirements in self-driving.

To further deal with localization error in practice, we can add pose noise as augmentation during training, and the results are listed in \cref{tab:pose}.
The robustness is further enhanced even for large pose errors, with some compromise of accuracy. 
Furthermore, as a future work, we can change convolution-based to attention-based BEV feature aggregation to alleviate misalignment caused by large localization errors.

\begin{table}[h]
  \centering
  \caption{Performance with the augmentation of adding pose noise on nuScenes~\cite{Caesar_2020_CVPR}, trained with 24 epochs. $\sigma_r$ is applied standard deviations of rotation noise.}
    \begin{tabular}{ccccccc}
    \toprule
    \makebox[2.0cm][c]{$\sigma_r$ (rad)} & \makebox[1.5cm][c]{0} & \makebox[1.5cm][c]{0.005} & \makebox[1.5cm][c]{0.01} & \makebox[1.5cm][c]{0.02} & \makebox[1.5cm][c]{0.05} & \makebox[1.5cm][c]{0.1} \\ 
    \midrule
    HRMapNet & 65.8 & 65.7 & 65.4 & 65.4 & 64.8 & 63.1 \\
    \bottomrule
    \end{tabular}%
  \label{tab:pose}%
\end{table}%

\section{Visualized Prediction Results}
\begin{figure}[ht]
    \centering
    \includegraphics[width=1\linewidth]{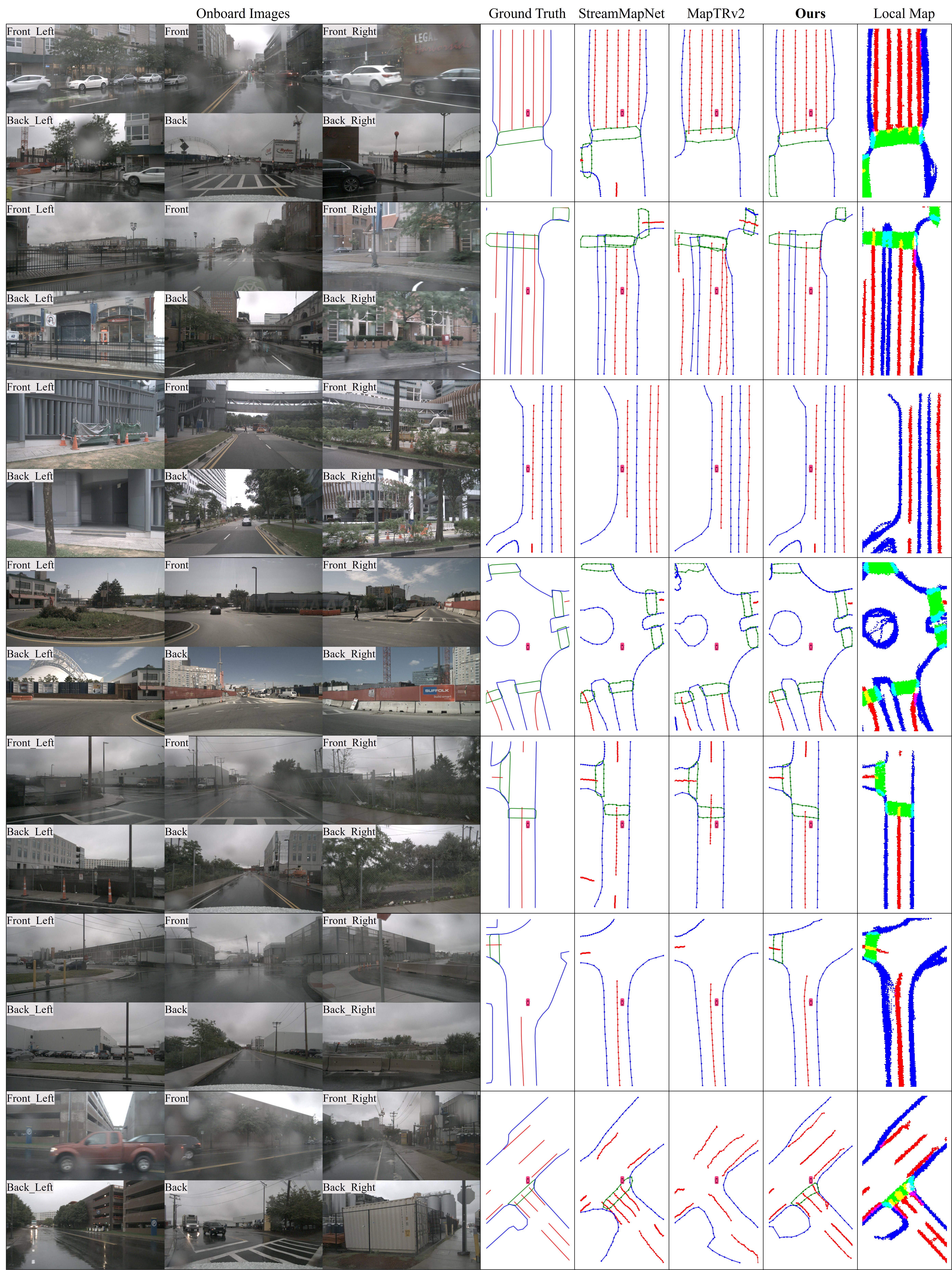}
    \caption{Visualized results.
    The last column is the retrieved local rasterized maps in {\name}.
    Lane divider, pedestrian crossing and road boundary are illustrated in red, green and blue respectively.}
    \label{fig:examples1}
\end{figure}

\begin{figure}[ht]
    \centering
    \includegraphics[width=1\linewidth]{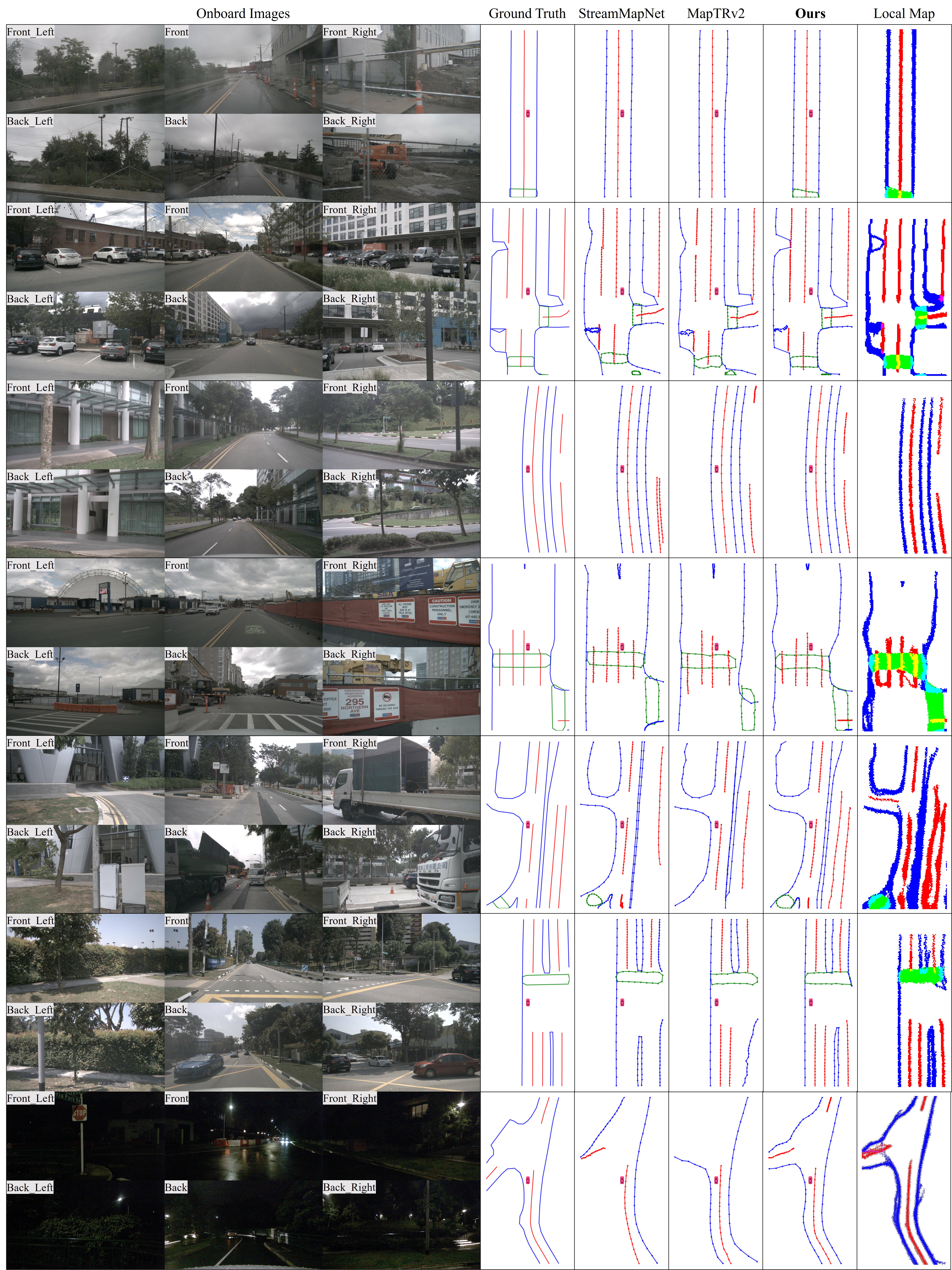}
    \caption{Visualized results. 
    The last column is the retrieved local rasterized maps in {\name}.
    Lane divider, pedestrian crossing and road boundary are illustrated in red, green and blue respectively.}
    \label{fig:examples2}
\end{figure}

\begin{figure}[h]
    \centering
    \includegraphics[width=1\linewidth]{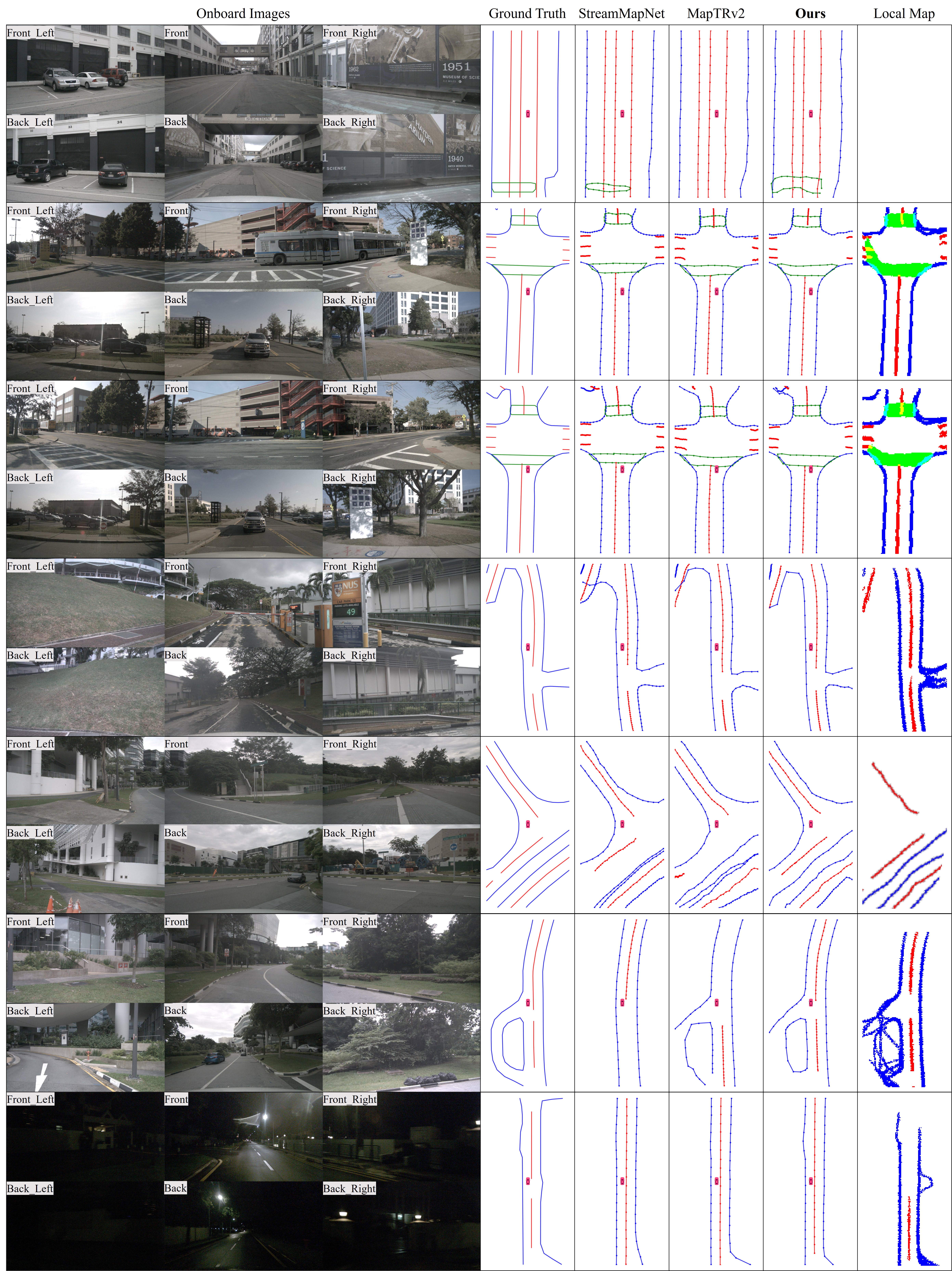}
    \caption{Visualized results.
    The last column is the retrieved local rasterized maps in {\name}.
    Lane divider, pedestrian crossing and road boundary are illustrated in red, green and blue respectively.}
    \label{fig:examples_map}
\end{figure}

In \cref{fig:examples1}, \cref{fig:examples2}, and \cref{fig:examples_map}, we provide more visualized comparison results. 
All three methods are trained with 24 epochs. Ours is based on MapTRv2. 
The last column are visualized retrieved local rasterized maps.
With the help of the local map, {\name} commonly achieves more accurate results, such as predicting map elements which are hard to recognize in images because of occlusion, bad weather or at long range. 

We evaluate constructed rasterized map using mean intersection over union (mIOU) and the result is mIOU=35.6 (ped:48.9, div:26.6, bou:31.5), which is not a high performance.
But our target is not to build and utilize a perfect HRMap. 
Although occasionally retrieved local maps are with noise or even error, {\name} still predicts good results from images.
Such examples are mainly illustrated in \cref{fig:examples_map}. 
The map noise or error is mainly from previous noisy predictions because of adverse weather, bad illumination or occlusions at turning or long range.
Since such noisy predictions take only a small part, these map noise would be removed gradually by multiple accurate predictions in long-term running. 
Specifically, for the example in the first row of \cref{fig:examples_map}, the global map is still empty since it is the first frame, but {\name} still provides accurate results. 
For the example in the second row of \cref{fig:examples_map}, the prediction results are not affected by the errors in the retrieved local map, caused by previous predictions.
And such map error is gradually removed by new accurate predictions, as illustrated in the third row.

\section{Visualized Global Rasterized Map}
\begin{figure}[ht]
    \centering
    \includegraphics[width=1\linewidth]{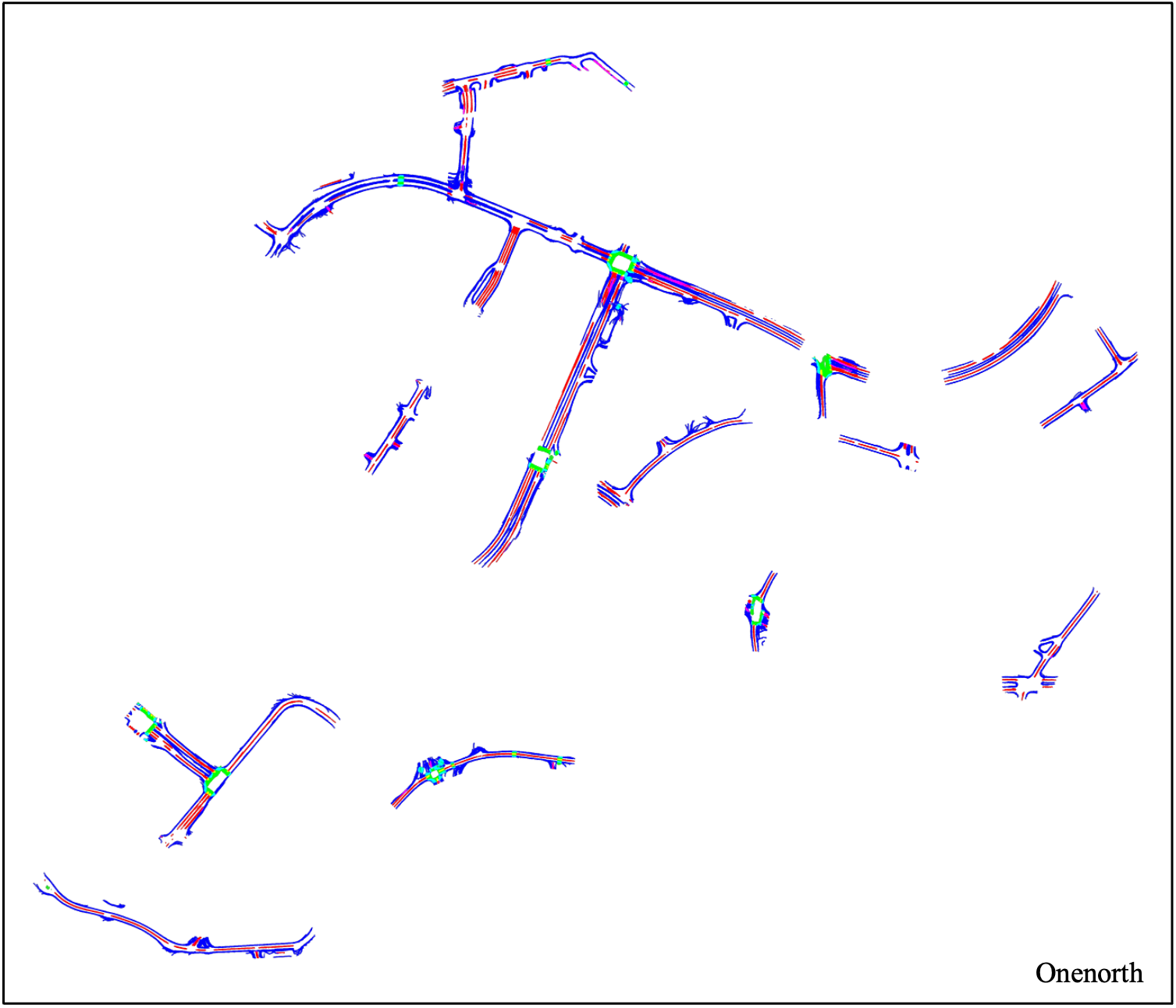}
    \includegraphics[width=1\linewidth]{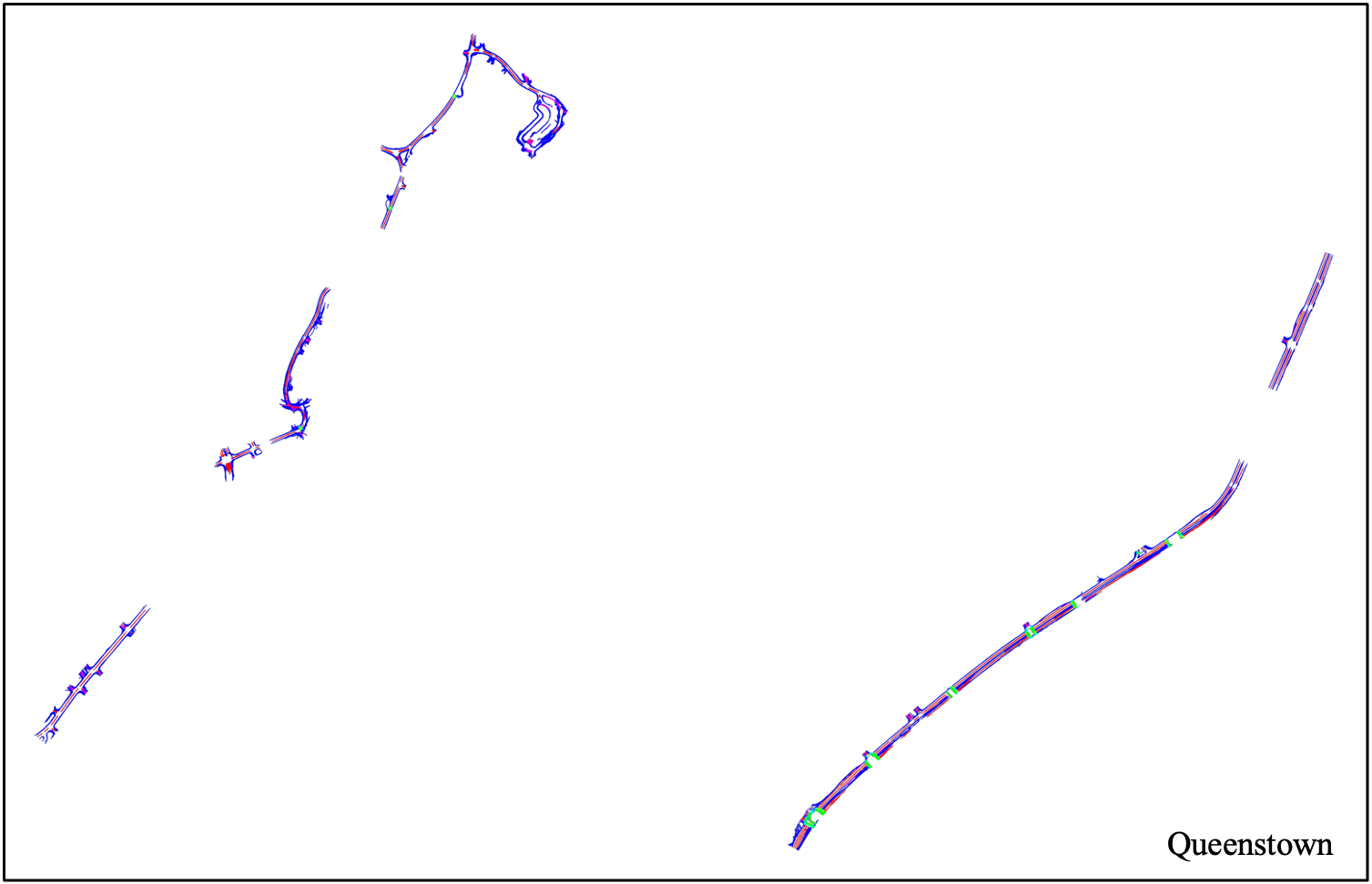}
    \caption{Visualized global rasterized maps.}
    \label{fig:maps}
\end{figure}

\begin{figure}[ht]
    \centering
    \includegraphics[width=1\linewidth]{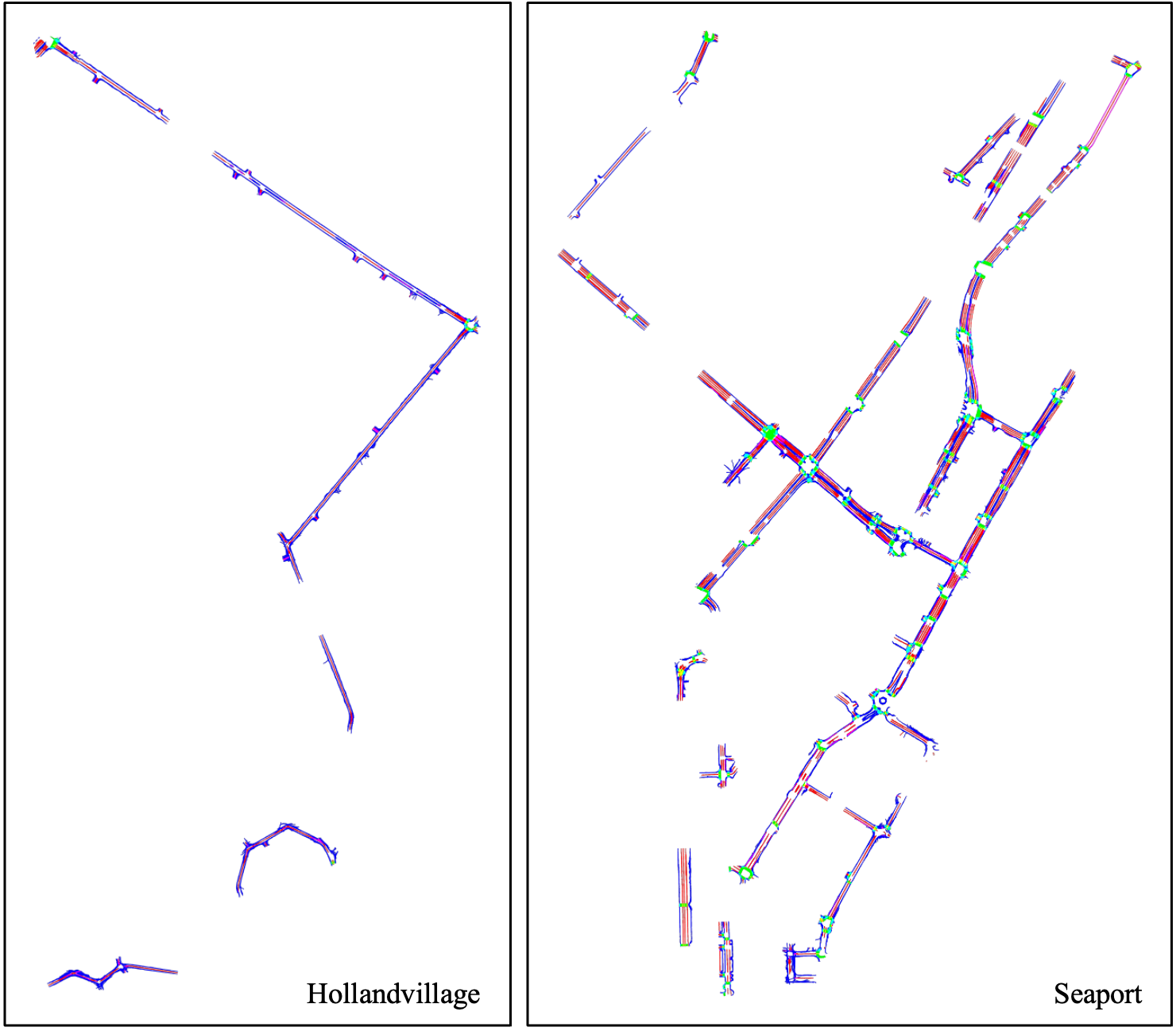}
    \caption{Visualized global rasterized maps.}
    \label{fig:maps2}
\end{figure}

We provide some visualized examples of the final global rasterized maps in \cref{fig:maps} and \cref{fig:maps2}. 
These global maps are constructed from empty ones, and updated gradually by prediction results of testing data.
We also provide a supplementary video for this process.


%% file: main.bbl
\begin{thebibliography}{10}
\providecommand{\url}[1]{\texttt{#1}}
\providecommand{\urlprefix}{URL }
\providecommand{\doi}[1]{https://doi.org/#1}

\bibitem{Caesar_2020_CVPR}
Caesar, H., Bankiti, V., Lang, A.H., Vora, S., Liong, V.E., Xu, Q., Krishnan, A., Pan, Y., Baldan, G., Beijbom, O.: {nuScenes: A multimodal dataset for autonomous driving}. In: IEEE/CVF Conference on Computer Vision and Pattern Recognition (CVPR) (2020)

\bibitem{Can_2021_ICCV}
Can, Y.B., Liniger, A., Paudel, D.P., {Van Gool}, L.: {Structured bird's-eye-view traffic scene understanding from onboard images}. In: IEEE/CVF International Conference on Computer Vision (ICCV). pp. 15661--15670 (2021)

\bibitem{Can_2022_CVPR}
Can, Y.B., Liniger, A., Paudel, D.P., {Van Gool}, L.: {Topology preserving local road network estimation from single onboard camera image}. In: IEEE/CVF Conference on Computer Vision and Pattern Recognition (CVPR). pp. 17263--17272 (2022)

\bibitem{carion2020end}
Carion, N., Massa, F., Synnaeve, G., Usunier, N., Kirillov, A., Zagoruyko, S.: {End-to-end object detection with transformers}. In: European Conference on Computer Vision (ECCV). pp. 213--229 (2020)

\bibitem{Chen2022}
Chen, S., Cheng, T., Wang, X., Meng, W., Zhang, Q., Liu, W.: Efficient and robust 2d-to-bev representation learning via geometry-guided kernel transformer. arXiv preprint arXiv:2206.04584  (2022)

\bibitem{Ding_2023_ICCV}
Ding, W., Qiao, L., Qiu, X., Zhang, C.: {PivotNet: Vectorized pivot learning for end-to-end HD map construction}. In: IEEE/CVF International Conference on Computer Vision (ICCV). pp. 3672--3682 (2023)

\bibitem{Feng_2022_CVPR}
Feng, Z., Guo, S., Tan, X., Xu, K., Wang, M., Ma, L.: {Rethinking efficient lane detection via curve modeling}. In: IEEE/CVF Conference on Computer Vision and Pattern Recognition (CVPR). pp. 17062--17070 (2022)

\bibitem{Gao_2020_CVPR}
Gao, J., Sun, C., Zhao, H., Shen, Y., Anguelov, D., Li, C., Schmid, C.: {VectorNet: Encoding HD maps and agent dynamics from vectorized representation}. In: IEEE/CVF Conference on Computer Vision and Pattern Recognition (CVPR) (2020)

\bibitem{Gao2023}
Gao, W., Fu, J., Jing, H., Zheng, N.: {Complementing onboard sensors with satellite map: A new perspective for HD map construction}. In: IEEE International Conference on Robotics and Automation (ICRA) (2024)

\bibitem{Gosala_2023_CVPR}
Gosala, N., Petek, K., Drews-Jr, P.L.J., Burgard, W., Valada, A.: Skyeye: Self-supervised bird's-eye-view semantic mapping using monocular frontal view images. In: IEEE/CVF Conference on Computer Vision and Pattern Recognition (CVPR). pp. 14901--14910 (2023)

\bibitem{grisetti2005improving}
Grisetti, G., Stachniss, C., Burgard, W.: Improving grid-based slam with rao-blackwellized particle filters by adaptive proposals and selective resampling. In: IEEE International Conference on Robotics and Automation (ICRA). pp. 2432--2437 (2005)

\bibitem{grisetti2007improved}
Grisetti, G., Stachniss, C., Burgard, W.: Improved techniques for grid mapping with rao-blackwellized particle filters. IEEE Transactions on Robotics  \textbf{23}(1),  34--46 (2007)

\bibitem{haklay2008openstreetmap}
Haklay, M., Weber, P.: Openstreetmap: User-generated street maps. IEEE Pervasive Computing  \textbf{7}(4),  12--18 (2008)

\bibitem{he2016deep}
He, K., Zhang, X., Ren, S., Sun, J.: Deep residual learning for image recognition. In: IEEE Conference on Computer Vision and Pattern Recognition (CVPR). pp. 770--778 (2016)

\bibitem{Huang2022}
Huang, J., Huang, G.: Bevpoolv2: A cutting-edge implementation of bevdet toward deployment. arXiv preprint arXiv:2211.17111  (2022)

\bibitem{Huang_2023_CVPR}
Huang, S., Shen, Z., Huang, Z., Ding, Z.h., Dai, J., Han, J., Wang, N., Liu, S.: Anchor3dlane: Learning to regress 3d anchors for monocular 3d lane detection. In: IEEE/CVF Conference on Computer Vision and Pattern Recognition (CVPR). pp. 17451--17460 (2023)

\bibitem{Jia_2023_CVPR}
Jia, D., Yuan, Y., He, H., Wu, X., Yu, H., Lin, W., Sun, L., Zhang, C., Hu, H.: {DETRs with hybrid matching}. In: IEEE/CVF Conference on Computer Vision and Pattern Recognition (CVPR). pp. 19702--19712 (2023)

\bibitem{jiang2024pmapnet}
Jiang, Z., Zhu, Z., Li, P., Gao, H.a., Yuan, T., Shi, Y., Zhao, H., Zhao, H.: P-mapnet: Far-seeing map generator enhanced by both sdmap and hdmap priors. arXiv preprint arXiv:2403.10521  (2024)

\bibitem{Lazarow_2022_CVPR}
Lazarow, J., Xu, W., Tu, Z.: {Instance segmentation with mask-supervised polygonal boundary transformers}. In: IEEE/CVF Conference on Computer Vision and Pattern Recognition (CVPR). pp. 4382--4391 (2022)

\bibitem{li2022hdmapnet}
Li, Q., Wang, Y., Wang, Y., Zhao, H.: {HDMapNet}: An online {HD} map construction and evaluation framework. In: IEEE International Conference on Robotics and Automation (ICRA). pp. 4628--4634 (2022)

\bibitem{li2024lanesegnet}
Li, T., Jia, P., Wang, B., Chen, L., JIANG, K., Yan, J., Li, H.: Lanesegnet: Map learning with lane segment perception for autonomous driving. In: International Conference on Learning Representations (ICLR) (2024)

\bibitem{li2022bevformer}
Li, Z., Wang, W., Li, H., Xie, E., Sima, C., Lu, T., Qiao, Y., Dai, J.: {Bevformer: Learning bird's-eye-view representation from multi-camera images via spatiotemporal transformers}. In: European Conference on Computer Vision (ECCV). pp. 1--18 (2022)

\bibitem{li2023fb}
Li, Z., Yu, Z., Wang, W., Anandkumar, A., Lu, T., Alvarez, J.M.: {FB-BEV: BEV representation from forward-backward view transformations}. In: IEEE/CVF International Conference on Computer Vision (ICCV). pp. 6919--6928 (2023)

\bibitem{lanegap}
Liao, B., Chen, S., Jiang, B., Cheng, T., Zhang, Q., Liu, W., Huang, C., Wang, X.: Lane graph as path: Continuity-preserving path-wise modeling for online lane graph construction. arXiv preprint arXiv:2303.08815  (2023)

\bibitem{Liao2023}
Liao, B., Chen, S., Wang, X., Cheng, T., Zhang, Q., Liu, W., Huang, C.: {MapTR: Structured modeling and learning for online vectorized HD map construction}. In: International Conference on Learning Representations (ICLR) (2023)

\bibitem{Liao2023-2}
Liao, B., Chen, S., Zhang, Y., Jiang, B., Zhang, Q., Liu, W., Huang, C., Wang, X.: Maptrv2: An end-to-end framework for online vectorized hd map construction. arXiv preprint arXiv:2308.05736  (2023)

\bibitem{pmlr-v202-liu23ax}
Liu, Y., Yuan, T., Wang, Y., Wang, Y., Zhao, H.: Vectormapnet: End-to-end vectorized hd map learning. In: International Conference on Machine Learning (ICML). pp. 22352--22369. PMLR (2023)

\bibitem{petr2022}
Liu, Y., Wang, T., Zhang, X., Sun, J.: {PETR: Position embedding transformation for multi-view 3D object detection}. In: European Conference on Computer Vision (ECCV). pp. 531--548 (2022)

\bibitem{Luo2023}
Luo, K.Z., Weng, X., Wang, Y., Wu, S., Li, J., Weinberger, K.Q., Wang, Y., Pavone, M.: Augmenting lane perception and topology understanding with standard definition navigation maps. arXiv preprint arXiv:2311.04079  (2023)

\bibitem{1087316}
Moravec, H., Elfes, A.: High resolution maps from wide angle sonar. In: IEEE International Conference on Robotics and Automation (ICRA). vol.~2, pp. 116--121 (1985)

\bibitem{philion2020lift}
Philion, J., Fidler, S.: {Lift, splat, shoot: Encoding images from arbitrary camera rigs by implicitly unprojecting to 3D}. In: European Conference on Computer Vision (ECCV) (2020)

\bibitem{Qiao2023}
{Qiao, Limeng and Ding, Wenjie and Qiu, Xi and Zhang}, C.: {End-to-end vectorized HD-map construction with piecewise bezier curve}. In: IEEE/CVF Conference on Computer Vision and Pattern Recognition (CVPR). pp. 13218--13228 (2023)

\bibitem{reid2019localization}
Reid, T.G., Houts, S.E., Cammarata, R., Mills, G., Agarwal, S., Vora, A., Pandey, G.: Localization requirements for autonomous vehicles. arXiv preprint arXiv:1906.01061  (2019)

\bibitem{Roddick_2020_CVPR}
Roddick, T., Cipolla, R.: Predicting semantic map representations from images using pyramid occupancy networks. In: IEEE/CVF Conference on Computer Vision and Pattern Recognition (CVPR) (2020)

\bibitem{9561375}
Schreiber, M., Belagiannis, V., Gläser, C., Dietmayer, K.: Dynamic occupancy grid mapping with recurrent neural networks. In: IEEE International Conference on Robotics and Automation (ICRA). pp. 6717--6724 (2021)

\bibitem{shan2018lego}
Shan, T., Englot, B.: {LeGO-LOAM}: Lightweight and ground-optimized lidar odometry and mapping on variable terrain. In: IEEE/RSJ International Conference on Intelligent Robots and Systems (IROS). pp. 4758--4765 (2018)

\bibitem{thrun2005probabilistic}
Thrun, S., Burgard, W., Fox, D.: Probabilistic Robotics. The MIT Press (2005)

\bibitem{vaswani2017attention}
Vaswani, A., Shazeer, N., Parmar, N., Uszkoreit, J., Jones, L., Gomez, A.N., Kaiser, L.u., Polosukhin, I.: Attention is all you need. In: Advances in Neural Information Processing Systems (NeurIPS) (2017)

\bibitem{wang2024openlane}
Wang, H., Li, T., Li, Y., Chen, L., Sima, C., Liu, Z., Wang, B., Jia, P., Wang, Y., Jiang, S., Others: {Openlane-v2: A topology reasoning benchmark for unified 3d hd mapping}. In: Advances in Neural Information Processing Systems (NeurIPS). vol.~36 (2024)

\bibitem{Wang_2023_CVPR}
Wang, R., Qin, J., Li, K., Li, Y., Cao, D., Xu, J.: {BEV-LaneDet: An efficient 3D lane detection based on virtual camera via key-points}. In: IEEE/CVF Conference on Computer Vision and Pattern Recognition (CVPR). pp. 1002--1011 (2023)

\bibitem{Wang_2023_ICCV}
Wang, S., Liu, Y., Wang, T., Li, Y., Zhang, X.: Exploring object-centric temporal modeling for efficient multi-view 3d object detection. In: IEEE/CVF International Conference on Computer Vision (ICCV). pp. 3621--3631 (2023)

\bibitem{Wang2024}
Wang, S., Jia, F., Liu, Y., Zhao, Y., Chen, Z., Wang, T., Zhang, C., Zhang, X., Zhao, F.: Stream query denoising for vectorized hd map construction. arXiv preprint arXiv:2401.09112  (2024)

\bibitem{Argoverse2}
Wilson, B., Qi, W., Agarwal, T., Lambert, J., Singh, J., Khandelwal, S., Pan, B., Kumar, R., Hartnett, A., Pontes, J.K., Ramanan, D., Carr, P., Hays, J.: Argoverse 2: Next generation datasets for self-driving perception and forecasting. In: Advances in Neural Information Processing Systems (NeurIPS) Datasets and Benchmarks Track (2021)

\bibitem{xie2023mv}
Xie, Z., Pang, Z., Wang, Y.X.: Mv-map: Offboard hd-map generation with multi-view consistency. In: IEEE/CVF International Conference on Computer Vision (ICCV). pp. 8658--8668 (2023)

\bibitem{Xiong_2023_CVPR}
Xiong, X., Liu, Y., Yuan, T., Wang, Y., Wang, Y., Zhao, H.: Neural map prior for autonomous driving. In: IEEE/CVF Conference on Computer Vision and Pattern Recognition (CVPR). pp. 17535--17544 (2023)

\bibitem{you2022hindsight}
You, Y., Luo, K.Z., Chen, X., Chen, J., Chao, W.L., Sun, W., Hariharan, B., Campbell, M., Weinberger, K.Q.: {Hindsight is 20/20: Leveraging past traversals to aid 3D perception}. In: International Conference on Learning Representations (ICLR) (2022)

\bibitem{yu2023scalablemap}
Yu, J., Zhang, Z., Xia, S., Sang, J.: Scalablemap: Scalable map learning for online long-range vectorized hd map construction. In: The 7th Conference on Robot Learning (CoRL). pp. 2429--2443. PMLR (2023)

\bibitem{Yuan_2024_WACV}
Yuan, T., Liu, Y., Wang, Y., Wang, Y., Zhao, H.: Streammapnet: Streaming mapping network for vectorized online hd map construction. In: IEEE/CVF Winter Conference on Applications of Computer Vision (WACV). pp. 7356--7365 (2024)

\bibitem{9197261}
Yue, Y., Zhao, C., Li, R., Yang, C., Zhang, J., Wen, M., Wang, Y., Wang, D.: A hierarchical framework for collaborative probabilistic semantic mapping. In: IEEE International Conference on Robotics and Automation (ICRA). pp. 9659--9665 (2020)

\bibitem{zhang2024online}
Zhang, G., Lin, J., Wu, S., song, y., Luo, Z., Xue, Y., Lu, S., Wang, Z.: Online map vectorization for autonomous driving: A rasterization perspective. In: Advances in Neural Information Processing Systems (NeurIPS). vol.~36, pp. 31865--31877 (2023)

\bibitem{zhang2014loam}
Zhang, J., Singh, S.: {LOAM}: Lidar odometry and mapping in real-time. In: Robotics: Science and Systems (RSS). vol.~2, pp.~1--9 (2014)

\bibitem{zhou2022cross}
Zhou, B., Kr{\"a}henb{\"u}hl, P.: Cross-view transformers for real-time map-view semantic segmentation. In: IEEE/CVF Conference on Computer Vision and Pattern Recognition (CVPR). pp. 13760--13769 (2022)

\end{thebibliography}
